\pdfoutput=1

\documentclass[11pt]{article}

\usepackage[final]{acl}

\usepackage{times}
\usepackage{latexsym}
\usepackage{booktabs} 
\usepackage{algorithm}
\usepackage{algpseudocode}
\usepackage{amsmath}
\usepackage{lipsum}
\usepackage{listings}
\usepackage{tcolorbox}
\usepackage{minted}

\usepackage[T1]{fontenc}

\usepackage[utf8]{inputenc}

\usepackage{microtype}

\usepackage{inconsolata}

\usepackage{graphicx}

\title{Large-Scale Aspect-Based Sentiment Analysis \\with Reasoning-Infused LLMs}

\author{Paweł Liskowski \\
Snowflake Inc. \\ Poznań, Poland \\\And
Krzysztof Jankowski \\
Snowflake Inc. \\ Warsaw, Poland \\ }

\begin{document}
    \maketitle
    \begin{abstract}
        We introduce Arctic-ABSA, a collection of powerful models for real-life aspect-based sentiment analysis (ABSA).
        Our models are tailored to commercial needs, trained on a large corpus of public data alongside carefully generated synthetic data, resulting in a dataset 20 times larger than SemEval14.
        We extend typical ABSA models by expanding the number of sentiment classes from the standard three (\textit{positive}, \textit{negative}, \textit{neutral}) to five, adding \textit{mixed} and \textit{unknown} classes, while also jointly predicting overall text sentiment and supporting multiple languages.
        We experiment with reasoning injection by fine-tuning on Chain-of-Thought (CoT) examples and introduce a novel reasoning pretraining technique for encoder-only models that significantly improves downstream fine-tuning and generalization.
        Our 395M-parameter encoder and 8B-parameter decoder achieve up to 10 percentage points higher accuracy than GPT-4o and Claude 3.5 Sonnet, while setting new state-of-the-art results on the SemEval14 benchmark. A single multilingual model maintains 87--91\% accuracy across six languages without degrading English performance.
        We release ABSA-mix, a large-scale benchmark aggregating 17 public ABSA datasets across 92 domains.
    \end{abstract}

    \section{Introduction}
    
    In recent years, aspect-based sentiment analysis (ABSA) has emerged as an essential technique in natural language processing. Unlike traditional sentiment analysis, which classifies the overall polarity of a text, ABSA identifies sentiments associated with specific aspects mentioned within the input. Formally, given a text $t$ and a set of target aspects $A = \{a_1, \ldots, a_n\}$, the goal is to predict a sentiment polarity $sp_i \in \{\textit{positive}, \textit{negative}, \textit{neutral}\}$ for each aspect $a_i$.
    
    While this formulation is standard in research, real-world applications demand more nuanced ABSA systems. First, customer feedback often contains mixed or ambiguous sentiments that do not fit neatly into standard polarity categories. Second, commercial use cases require multi-level analysis that combines sentence-level evaluation with aspect-level granularity—capturing both overall user sentiment and detailed feedback on specific features. Third, global deployment necessitates multilingual support, a challenge compounded by linguistic diversity and scarce annotated data in many languages. Meeting industry-level quality standards further requires scalable systems that generalize across diverse domains.

    Large language models (LLMs) offer compelling advantages for addressing these challenges. Recent advances have yielded models with strong capabilities in contextual understanding and semantic representation—both critical for effective ABSA. By leveraging learned embedding spaces and reasoning abilities, LLMs can interpret complex sentiment expressions, including implicit opinions and figurative language that often confound traditional methods. Moreover, LLMs enable the development of synthetic data pipelines, facilitating the generation of diverse, high-quality training samples and reducing reliance on costly manual annotation.
    
    Motivated by these opportunities, we introduce Arctic-ABSA, a suite of models tailored for commercial-scale ABSA deployment. Our key contributions are:
    \begin{itemize}
        \item A suite of fine-tuned encoder-only and decoder-only LLMs, including thinking-mode variants trained on chain-of-thought (CoT) reasoning examples. Our 8B decoder sets new state-of-the-art on SemEval14 (91.8\% on restaurants, 87.2\% on laptops) while achieving 93\% accuracy on ABSA-mix---10 percentage points above Claude 3.5 Sonnet. Even our 395M encoder outperforms 405B-parameter models on most benchmarks. A single multilingual encoder achieves 87--91\% accuracy across six languages, demonstrating strong cross-lingual generalization.
        \item A novel \textit{upside-down} synthetic data generation method that inverts the conventional ABSA workflow: rather than extracting sentiments from text, we generate text conditioned on predefined aspect-sentiment pairs, enabling scalable production of diverse, domain-spanning training samples.
        \item An efficient pretraining technique that embeds reasoning capabilities into encoder-only LLMs, significantly improving training efficiency and downstream performance.
        \item ABSA-mix, a large-scale benchmark dataset combining 17 public ABSA sources across 92 domains, which we release to facilitate future research.
    \end{itemize}

    \begin{figure*}[ht!]

        \includegraphics[trim=0 21.6cm 0 0, width=\linewidth]{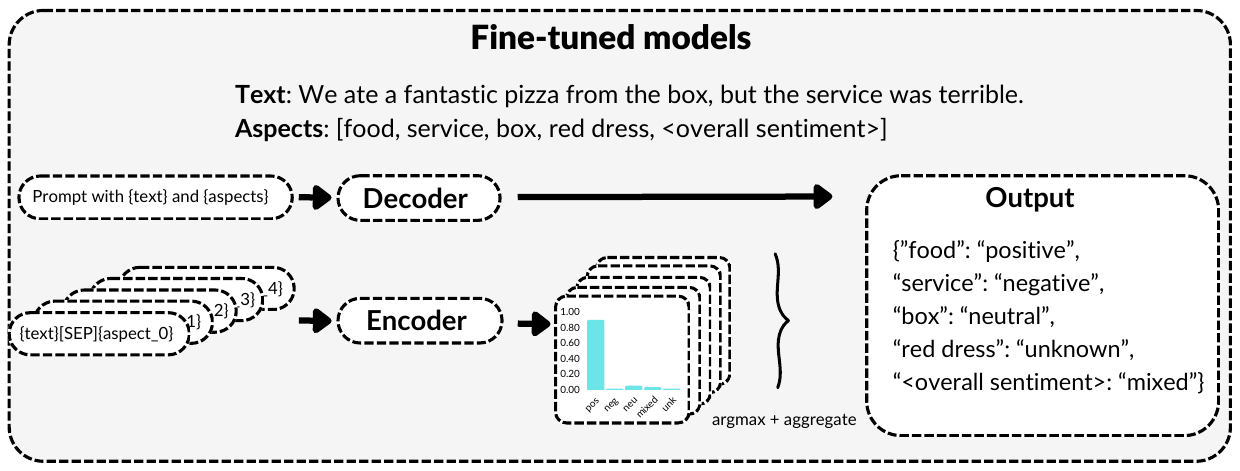}
        \caption {Architecture of fine-tuned models (decoder and Encoder) alongside example input and output.
        The decoder-only model directly outputs a JSON response for all aspects, whereas the encoder-only model requires as many forward passes as there are aspects.}
        \label{fig:models}
    \end{figure*}

    \section{Related work}
    
    \paragraph{Model Architectures for ABSA.}
    Aspect-based sentiment analysis has been extensively studied, with a variety of architectures proposed. Early approaches centered on encoder-only models based on BERT \citep{devlin2019bertpretrainingdeepbidirectional} and its variants, enhanced with task-specific architectural modifications such as local context focus \citep{Li2021LearningIS}, domain adaptation \citep{rietzler-etal-2020-adapt}, and relational graph attention \citep{wang-etal-2020-relational}. Encoder-decoder models also gained traction, with fine-tuned T5 \citep{raffel2023exploringlimitstransferlearning} and T\textit{k}-Instruct adopted in several studies \citep{ouyang2023aspectbasedsentimentanalysisexplicit, scaria_instructabsa_2023}. With the rise of LLMs, researchers began exploring decoder-only models in both zero-shot and fine-tuned settings \citep{simmering2023largelanguagemodelsaspectbased, smid-etal-2024-llama}.

    \paragraph{Reasoning for Sentiment Analysis.}
    Recent advances in reasoning-augmented LLMs have shown promise for complex NLP tasks. Chain-of-Thought (CoT) prompting \citep{wei2023chainofthoughtpromptingelicitsreasoning} enables models to decompose problems into intermediate steps, improving performance on tasks requiring multi-step inference. In ABSA, reasoning approaches have been applied through explicit rationale generation \citep{fei_reasoning_2023} and retrieval-augmented reasoning \citep{lai_rvisa_2024}. The Self-Taught Reasoner (STaR) framework \citep{zelikman2022starbootstrappingreasoningreasoning} demonstrated that models can bootstrap their own reasoning capabilities through iterative refinement. Our work extends these ideas by injecting reasoning directly into encoder pretraining, a direction that remains underexplored.

    \paragraph{Synthetic Data for Sentiment Analysis.}
    Data augmentation and synthetic data generation have become essential for improving model robustness, particularly in low-resource settings. Traditional approaches include back-translation, synonym replacement, and paraphrasing \citep{feng-etal-2021-survey}. More recently, LLMs have been used to generate training data at scale \citep{ye-etal-2022-zerogen, wang_self-instruct_2023}. For sentiment analysis specifically, prior work has explored generating labeled examples through prompting \citep{li-etal-2023-synthetic}. Our \textit{upside-down} approach differs by inverting the typical workflow: instead of generating text and then labeling it, we specify aspect-sentiment pairs first and generate text that reflects those constraints, enabling precise control over data distribution.

    \paragraph{Multilingual ABSA.}
    Cross-lingual sentiment analysis presents unique challenges due to linguistic diversity and limited annotated resources in non-English languages. Early approaches relied on machine translation or cross-lingual embeddings \citep{barnes-etal-2018-bilingual}. Recent multilingual transformers such as mBERT and XLM-R \citep{conneau-etal-2020-unsupervised} have enabled zero-shot cross-lingual transfer, though performance gaps remain between high- and low-resource languages. Our work demonstrates that a single encoder trained on translated data can achieve consistently strong performance across multiple languages without sacrificing English accuracy.
    
    Our approach builds on these advances by combining reasoning-augmented training with both encoder-only and decoder-only architectures, while introducing novel techniques for synthetic data generation and reasoning injection in encoders.

    \section{Fine-tuning LLMs for ABSA} \label{sec:methodology}
    
    Our choice of architectures is motivated by recent trends in both decoder-only and encoder-only models. The availability of powerful decoder-only LLMs \citep{qwen2025qwen25technicalreport, grattafiori_llama_2024} and advances in reasoning techniques \citep{deepseek-ai_deepseek-r1_2025} make them attractive for ABSA. At the same time, the emergence of efficient, high-performing encoders \citep{warner_smarter_2024, breton_neobert_2025, boizard2025eurobertscalingmultilingualencoders} and the historically strong performance of BERT-like models on ABSA tasks motivate the inclusion of an encoder-only variant.
    
    \subsection{Decoder-only LLMs} \label{dec-only-llm}
    We fine-tune LLaMA3.1-8B \citep{grattafiori_llama_2024} for ABSA using an instruction-tuning approach.

    Our method formulates ABSA as a sequence-to-sequence task. The input is a text passage combined with explicit instructions for aspect extraction and sentiment classification. The model is trained to generate structured outputs in JSON format, encoding both the extracted aspects and their associated sentiment polarities, along with an overall sentiment assessment of the passage.

    We employ Low-Rank Adaptation (LoRA) \cite{hu2022lora} for parameter-efficient fine-tuning, which adds trainable low-rank matrices to the attention layers of the model. Our LoRA configuration uses a rank of 64 and an alpha of 128, applying adaptations to the query ($W_q$) and value ($W_v$) projection matrices in the self-attention layers of transformer blocks. 

    The training objective is causal language modeling, implemented via next-token prediction. We use a token-level cross-entropy loss, applying a loss mask that excludes input (prompt) tokens and computes the loss only on target (completion) tokens. Optimization is performed using the AdamW algorithm \citep{loshchilov2019decoupledweightdecayregularization}, with an effective batch size of $8$, a learning rate of $5 \times 10^{-5}$, cosine annealing for learning rate scheduling, $1000$ warm-up steps, and no weight decay.
    
    \subsection{Encoder-only LLMs}
    Decoder-only models, which typically consist of billions of parameters, are often computationally demanding to train and deploy, making them prohibitively expensive in scenarios where efficiency is a key consideration. To address this constraint, we explore ModernBERT \citep{warner_smarter_2024}, a lightweight, BERT-style encoder-only architecture that offers a strong balance between performance and computational cost.

    We adapt ModernBERT-large (395M parameters) to the ABSA task by adding a task-specific classification head consisting of a linear layer that maps the final-layer \texttt{[CLS]} token representation to logits over five sentiment classes. Inputs to the model are structured as:
    \begin{center}
    \texttt{[CLS] text [SEP] aspect [SEP]}
    \end{center}
    The \texttt{[SEP]} token delineates the main text from the target aspect term and encourages the model to establish cross-attention between the term and relevant context within the text. For instances containing multiple aspects, we create separate input sequences for each aspect-text pair and process them independently.
        
    Sentiment classification is based on the final-layer representation of the \texttt{[CLS]} token, passed through a fully connected layer to generate class logits. Formally:
    $$h_{CLS} = \text{BERT}(input)_{[CLS]}$$
    $$y = \text{softmax}(Wh_{CLS} + b)$$
    
    We train the model using a cross-entropy loss function. Given the predicted probability distribution $y$ and the one-hot ground truth label $y^*$, the loss is computed as:
    $$\mathcal{L} = -\sum_{c=1}^{C} y^*_c \log(y_c)$$
    where $C$ is the number of sentiment classes.

    Fine-tuning is performed using the AdamW optimizer with hyperparameters selected based on preliminary experimentation: a learning rate of $3 \times 10^{-5}$, batch size of $64$, and training for $3$ epochs. We apply a linear learning rate decay schedule (no warm-up) and a weight decay of $0.01$ to regularize training and reduce overfitting. The optimizer updates both the pre-trained ModernBERT parameters and the added classification head. Figure\ \ref{fig:models} illustrates the fine-tuning process.

    \subsection{Reasoning injection}\label{reason-inject}
    Inspired by the growing effectiveness of reasoning-augmented LLMs \citep{deepseek-ai_deepseek-r1_2025, openai2024openaio1card} and their application to ABSA \citep{lai_rvisa_2024, fei_reasoning_2023}, we develop variants of our models that explicitly incorporate reasoning into the training process.
    
    To enable aspect-level reasoning, we augment the training dataset by generating reasoning chains for each sample. Our methodology follows a two-phase process adapted from the Self-Taught Reasoner (STaR) framework \citep{zelikman2022starbootstrappingreasoningreasoning}:
    \begin{itemize}
        \item \textbf{Initial Reasoning Generation}. For each training sample containing a list of aspects, we prompt an LLM to produce a reasoning chain using few-shot examples\ \citep{brown2020languagemodelsfewshotlearners}  and CoT\ \citep{wei2023chainofthoughtpromptingelicitsreasoning} prompting techniques.

        \item \textbf{Reasoning Refinement}. We verify the model’s answer against the ground truth. If the output is incorrect, we prompt an LLM with both the original reasoning and the correct answer, instructing it to revise the reasoning to reach the correct conclusion.
    \end{itemize}

    To ensure our training data contains only high-quality reasoning chains that effectively demonstrate the path to correct ABSA outcomes, we curate our datasets by discarding samples in which the refinement process either failed to produce the correct answer or the LLM refused to generate output.

    We then implement a two-phase training process comprising pretraining and fine-tuning stages. During pretraining, we inject reasoning capabilities in a supervised manner by training the model to predict reasoning chains. In the fine-tuning phase, we mask out the reasoning chains and train the model solely on predicting the correct aspect-sentiment pairs. The specifics of the pretraining setup differ across model architectures, and we detail these differences in the following sections.

    \subsubsection{Decoder-only reasoning training}

    For decoder-only architectures, reasoning injection during pretraining can be implemented through either causal language modeling or supervised training. We adopt the latter by training the model to predict reasoning chains before making final predictions. Specifically, we extend our structured output format to include a leading key-value pair with the key \textit{<thoughts>}, which precedes the aspect-sentiment pairs.
    
    This approach requires the model to first articulate its reasoning process—analyzing relationships between aspects and their contextual sentiment indicators—before generating the final structured output. By forcing the model to predict this reasoning chain as part of its target sequence, we teach it to develop systematic analytical thinking patterns. Our methodology aligns with chain-of-thought internalization research \citep{deng_implicit_2023, hao_training_2024}, where the goal is to develop models that incorporate the benefits of explicit reasoning while producing only the final answer at inference time.

    \subsubsection{Encoder-only thinking mode}
    To inject reasoning capabilities into encoder-only models, we utilize training data enriched with generated reasoning chains for each sample (cf. Section \ref{reason-inject}) and structure training examples using the following template:
    
    \begin{flushleft}
    \texttt{<input text>: \{\}} \\
    \texttt{<input aspects>: \{\}} \\
    \texttt{<intermediate reasoning>: \{\}} \\
    \texttt{<output sentiments for aspects>: \{\}}
    \end{flushleft}
    
    During pretraining, we apply Masked Language Modeling (MLM) \citep{devlin_bert_2019} to train the model to capture contextual dependencies within the input. Crucially, the structured template serves as a scaffold, guiding the model in understanding the relationships between the input text, aspects, reasoning chains, and final sentiment classifications. In the subsequent fine-tuning phase, we first attach a task-specific classification head and then train the model to predict aspect sentiments based on the learned representations in a supervised manner.

    A key innovation in our approach is the \textit{targeted masking} strategy, which operates as follows: first, a random 30\% of tokens are selected for masking, ensuring that all sentiment classification tokens are included in this selection. Then, conventional MLM masking is applied, where 80\% of the selected tokens are masked, 10\% are replaced with random tokens, and the remaining 10\% remain unchanged. Unlike random masking, the targeted strategy forces the model to always predict sentiment tokens during training. In this way, it creates a stronger learning signal for the sentiment classification task and forces the model to use reasoning chains to predict the masked tokens.

    To optimize computational efficiency during pretraining, we implement an unpadding technique, where multiple structured examples are concatenated and we use a \texttt{[SEP]} token as a delimiter \citep{zeng2022boostingdistributedtrainingperformance, portes2024mosaicbertbidirectionalencoderoptimized, zhang-etal-2024-mgte}. Each batch contains sequences of up to $L_{max}$, fitting as many complete samples as possible without exceeding the limit.

    In our experiments, we pretrain models for $4$ epochs with a batch size of $32$ and a learning rate of $1 \times 10^{-5}$.
    The optimization is conducted using the AdamW optimizer with a linear learning rate scheduler.

    \begin{table*}[ht!]
        \centering
        \caption{Accuracy of popular LLMs (proprietary and open-source) and Arctic-ABSA models on ABSA-mix, overall sentiments dataset, SemEval14 (restaurants + laptops), FABSA, and SENTFIN. The best result is \textbf{bolded}, and the second best is \underline{underlined}.}
        \begin{tabular}{lrrrrrr}
            \toprule
            \textbf{Model Name} & \textbf{ABSA-mix}   & \textbf{Overalls}   & \textbf{Lapt14}     & \textbf{Rest14} & \textbf{Fabsa} & \textbf{Sentfin} \\
            \midrule
            
        \multicolumn{7}{c}{\small\textit{Closed-source models}} \\
        \cmidrule(lr){1-7} 
        
            GPT-4o                & 82.15\%             & 73.00\%             & 83.36\%             & 87.19\%             & 77.35\%             & 81.20\%             \\
            Claude 3.5 Sonnet   & 83.65\%             & 71.83\%             & 83.36\%             & 87.90\%             & 82.01\%             & 84.02\%             \\
            \cmidrule(lr){1-7}
            \multicolumn{7}{c}{\small\textit{Open-source models}} \\
            \cmidrule(lr){1-7}
            Mistral Large 2     & 82.77\%             & 76.50\%             & 80.98\%             & 87.54\%             & 81.22\%             & 81.20\%             \\
            Llama3.1-405b       & 83.08\%             & 73.67\%             & 82.41\%             & \textbf{89.87}\%    & 82.83\%             & 83.61\%             \\
            Llama3.1-70b        & 81.03\%             & 72.33\%             & 78.92\%             & 85.30\%             & 76.42\%             & 76.86\%             \\
            Llama3.1-8b         & 76.65\%             & 79.17\%             & 73.22\%             & 81.09\%             & 81.94\%             & 62.88\%             \\
            Llama3.2-3b         & 70.22\%             & 79.50\%             & 73.38\%             & 80.47\%             & 91.29\%             & 61.57\%             \\
            \cmidrule(lr){1-7}
            \multicolumn{7}{c}{\small\textit{Arctic-ABSA models}} \\
            \cmidrule(lr){1-7}
            Encoder             & 91.28\% & 88.17\% & 83.99\%             & \underline{89.34}\% & 96.45\% & 90.98\% \\
            Encoder-thinking    & 91.24\%             & 87.67\%             & \underline{85.10}\% & 89.16\%             & 96.42\%             & \underline{91.12}\% \\
            Decoder             & \textbf{93.03}\%    & \underline{89.77}\%    & \textbf{86.05}\%    & 88.17\%             & \underline{96.59}\% & \textbf{91.53}\% \\
            Decoder-thinking    & \underline{92.99}\%               & \textbf{90.00}\%               & \underline{85.10}\%               & 87.19\%               & \textbf{97.17}\%               & \underline{91.12}\%               \\

            \bottomrule
        \end{tabular}
        \label{tab:model_performance}
    \end{table*}

    \section{Dataset}\label{sec:dataset}
    Aspect-based sentiment analysis models have traditionally been trained and evaluated on small, domain-specific datasets, limiting their applicability in real-world, cross-domain scenarios. In response, we constructed a large-scale dataset, \textbf{ABSA-public}, by aggregating 17 publicly available ABSA datasets. A complete list of these sources is provided in Table~\ref{tab:dataset_sources} in the Appendix.

    Combining multiple datasets requires careful preprocessing to avoid introducing duplicates or test set leakage. To ensure data quality, we performed thorough cleaning. Duplicate samples, identified by matching text values, were merged by unifying their associated aspect lists. In cases where the same text had conflicting sentiment labels for the same aspect (e.g., positive and negative), we assigned a \textit{mixed} label to reflect the ambiguity. Additionally, we conducted rigorous checks to identify and remove any data leaks from the training and validation sets.

    \subsection{Upside-Down Synthetic Data Generation}
    \label{sec:upside-down}

    The performance of ABSA models relies heavily on high-quality labeled training samples, which are often costly and labor-intensive to produce. Moreover, developing a robust model capable of handling real-world scenarios requires training data that covers diverse domains and extends beyond the standard polarity categories.

    To overcome limitations in data availability, we introduce a novel data generation strategy called \textit{upside-down synthetic data generation} (UPSD). The proposed approach inverts the traditional ABSA workflow: rather than analyzing existing texts to identify aspect-level sentiments, we begin with \textit{predefined} aspect-sentiment pairs and use an LLM to generate coherent and realistic texts that reflect those sentiments.
    
    UPSD leverages the generative capabilities of LLMs to produce controllable, sentiment-aligned text. For example, given a set of inputs such as:
    \begin{flushleft}
    \texttt{battery life: positive} \\
    \texttt{price: negative} \\
    \texttt{customer service: neutral}
    \end{flushleft}
    the LLM is prompted to compose a text that naturally embeds these sentiment polarities in appropriate linguistic context. The UPSD workflow allows practitioners to generate balanced data distributions, target underrepresented aspect-sentiment combinations, or simulate realistic scenarios tailored to specific application domains. Moreover, it provides a flexible framework for producing data with varying tone, formality, or style, further enhancing the adaptability of ABSA systems trained on such data.

    The UPSD process involves three key steps: (1) defining a set of aspects and their corresponding target sentiment values, (2) constructing a prompt that communicates these specifications to the LLM, including optional constraints such as tone or length, and (3) generating text in which the specified sentiments are linguistically embedded and clearly associated with the corresponding aspects. 
    
    \subsection{Generating Mixed-Sentiment Data}

    Using the upside-down approach, we generated a synthetic dataset, \textbf{ABSA-synth}, to complement ABSA-public and address specific gaps—namely, the lack of \textit{mixed} and \textit{unknown} sentiment classes.

    To ensure both quality and diversity in ABSA-synth, we designed a multi-stage generation pipeline:
    \begin{enumerate}
        \item \textbf{Domain diversification}: Generate a wide range of topical categories across domains (e.g., electronics, hospitality, customer support).
        \item \textbf{Aspect enrichment}: For each category, select relevant aspects and augment them with guiding keywords to encourage variety in expression.
        \item \textbf{Controlled generation}: Randomly sample aspect-sentiment pairs, including \texttt{mixed} and \texttt{unknown} labels, and compose prompts for the LLM.
        \item \textbf{Stylistic variation}: Apply handcrafted \textit{persona} templates to the prompts to induce diverse styles, tones, and perspectives~\citep{ge_scaling_2024}.
    \end{enumerate}

    The above pipeline is an instantiation of the UPSD framework, which allows us to generate text samples that are sentimentally accurate while exhibiting broad lexical and stylistic diversity. The resulting dataset provides valuable coverage of edge-case sentiment classes and significantly enriches the training signal for ABSA models. A visual overview of the generation process is provided in Figure~\ref{fig:synthetic_data} in the Appendix.

    \subsection{Introducing the Unknown Class}

    Creating samples with the \textit{unknown} sentiment is comparatively straightforward, as it does not require generating new text but rather assigning an aspect that is absent from the existing text. To introduce this class, we augment $25\%$ of the text samples in our dataset by adding randomly selected aspects labeled with \textit{unknown} sentiment.

    To ensure the correctness of these assignments, we employ LLM-as-a-judge~\citep{li2025generationjudgmentopportunitieschallenges, gu2025surveyllmasajudge}. Specifically, the LLM is tasked with verifying that the selected aspect is not implicitly or explicitly mentioned in the given text. While naive random sampling from the set of non-explicit aspects might suffice in principle, it risks introducing false negatives by including aspects that are referenced implicitly. 
    The full procedure is described in Algorithm~\ref{alg:unknown_aspect} in the Appendix.

    \subsection{SemEval synthetic data}
    When scaling training data for real-world ABSA applications, we observed that model performance on popular benchmark datasets such as SemEval-2014 can degrade due to subtle differences in data distributions. This issue is particularly pronounced for the \textit{neutral} sentiment class, which is inconsistently annotated across datasets. In many cases, annotators vary in how they interpret weak or ambiguous expressions of sentiment, leading to a blurred boundary between \textit{neutral}, \textit{positive}, and \textit{negative} labels.

    To mitigate this problem and improve performance on the SemEval-2014 dataset, we applied \textit{dataset upscaling}. This data augmentation strategy generates synthetic samples closely aligned with the linguistic and structural characteristics of the original dataset, while increasing the representation of the \textit{neutral} sentiment class.
    
    Our upscaling procedure works as follows: we partition the original SemEval-2014 dataset into random batches of 10 samples and prompt an LLM to generate a new text sample for each batch. The prompt is designed to encourage similarity in style, topic, and structure, while explicitly prioritizing the inclusion of \textit{neutral} sentiment where appropriate. The process is repeated iteratively until the number of generated samples approaches the size of the original dataset. The resulting synthetic dataset, which we refer to as \textbf{SemEval14-synth}, is then post-processed through a cleaning and deduplication step to ensure quality and prevent redundancy.
    
    The complete procedure is described in Algorithm~\ref{alg:semeval_synth} in the Appendix.

    \subsection{Large-scale dataset}
    Our final dataset, \textbf{ABSA-mix}, combines ABSA-public and ABSA-synth, with \textit{unknown} sentiment labels added to 25\% of text samples. For each augmented sample, the number of \textit{unknown} aspects is drawn from a uniform distribution $\mathcal{U}(2,4)$. To further enhance model performance on the SemEval-2014 benchmark, we also incorporate the SemEval14-synth dataset into ABSA-mix. A summary of dataset statistics is provided in Table~\ref{tab:dataset_stats} in the Appendix.  

    \section{Results}
    We evaluate our models on a variety of datasets including our ABSA-mix dataset, Overalls ($600$ uniformly sampled datapoints from a concatenation of IMDB\ \citep{maas_learning_2011}, SST2\ \citep{sst2} and MSAD\ \citep{msad} datasets), SemEval14\ \citep{pontiki_semeval-2014_2014} (restaurants + laptops), FABSA\ \citep{kontonatsios_fabsa_2023} and SENTFIN\ \citep{sinha_sentfin_2022}.

    To ensure a fair comparison, we compare our models to popular LLMs trained on diverse corpora rather than specialized ABSA datasets.
    The results are summarized in Table\ \ref{tab:model_performance}.
    Our models significantly outperform even LLMs that are orders of magnitude larger.

    \subsection{SemEval14 performance}
    As SemEval14 remains a popular benchmark for ABSA models, we also evaluate our models on this dataset.
    In this scenario, we train our models only on the SemEval14 dataset and compare them to the state-of-the-art ABSA models also trained on the SemEval14 dataset.
    The baselines include a popular encoder-decoder model - Instruct-ABSA-2\ \citep{scaria_instructabsa_2023}, an encoder-only model LSA{\textsubscript{\textit{E}}}-X-DeBERTa\ \citep{yang2024lsamodelingaspectsentiment} pretrained on a larger text corpus than typical ABSA models, and a reasoning-based model RVISA\ \citep{lai_rvisa_2024} with various Flan-T5\ \citep{flan_t5} backbone sizes.

    We also report results for models trained on SemEval14 augmented with SemEval14-synth, which we found generally improves performance for smaller encoder-only models. Adding this synthetic data did not significantly improve decoder-only model performance, so we omit those results for brevity.

    \begin{table}[h!]
        \centering
        \begin{tabular}{lrr}
            \toprule
            \textbf{Model}                                          & \textbf{Rest14} & \textbf{Lapt14} \\
            \midrule
            Instruct-ABSA-2 \textit{200M}                           & 88.03\%         & 83.60\%         \\
            LSA{\textsubscript{\textit{E}}}-X-DeBERTa \textit{418M} & 90.98\%         & 86.46\%         \\
            Flan-T5+RVISA{\textsubscript{\textit{g}}} \textit{250M} & 89.29\%         & 81.82\%         \\
            Flan-T5+RVISA{\textsubscript{\textit{g}}} \textit{11B}  & 91.52\%         & 86.68\%         \\
            \cmidrule(lr){1-3}
            \multicolumn{3}{c}{\small\textit{Arctic-ABSA models}} \\

            \cmidrule(lr){1-3}
            Encoder \textit{395M}                                   & 88.35\%         & 83.83\%         \\
            \hspace*{1em} +synth                                    & 89.34\%         & 83.20\%         \\
            Encoder-thinking \textit{395M}                          & 88.71\%         & 83.83\%         \\
            \hspace*{1em} +synth                                    & 90.05\%         & 85.42\%         \\
            Decoder \textit{8B}                                     & \textbf{91.76}\%         & \textbf{87.16}\%         \\

            \bottomrule
        \end{tabular}
        \caption{Arctic-ABSA models accuracy compared to state-of-the-art ABSA models trained only on SemEval14 dataset (restaurants + laptops) with $3$ sentiments as output.}
        \label{tab:semeval_performance}
    \end{table}

    When trained on the SemEval14 dataset, our models are on par with the state-of-the-art models.
    Adding thinking mode boosts the performance of the encoder-only model in this lower-resource setting.
    Also, adding the synthetic data SemEval14-synth improves the performance of our encoder-only models and is a promising direction for enlarging the training data in a low-data regime for small models.

    \begin{figure}[h!]
        \includegraphics[width=\linewidth]{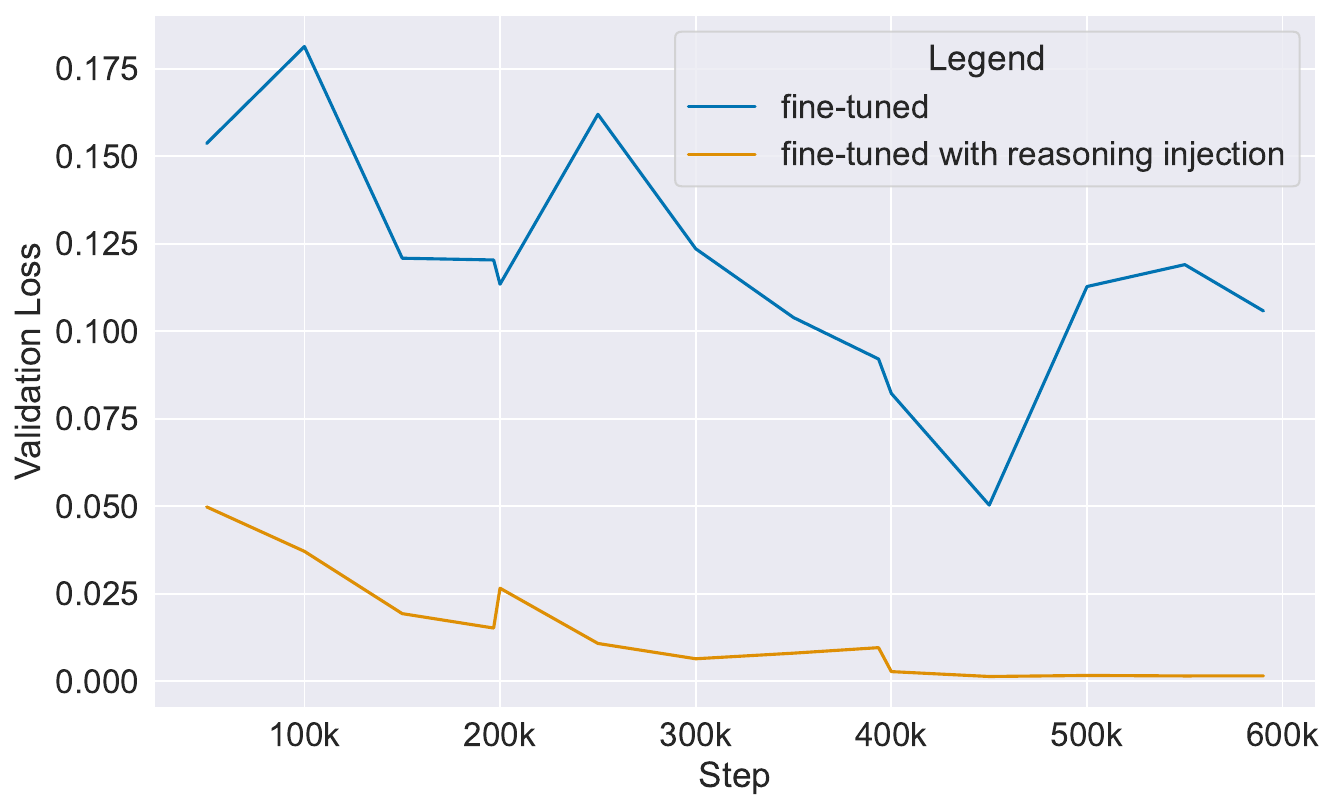}
        \caption {Comparison of validation losses of ModernBERT fine-tuned on ABSA-mix dataset and the variant with reasoning injection pretraining.
        Reasoning injection significantly improves the validation loss and leads to better generalization properties.}
        \label{fig:encoder_training}
    \end{figure}

    \subsection{Reasoning injection}
    
    While reasoning injection in encoder-only models may not yield the highest gains in accuracy alone, it offers significant benefits in training efficiency and generalization capability. As shown in Figure~\ref{fig:encoder_training}, the model trained with reasoning injection exhibits a substantially lower and smoother validation loss compared to its baseline counterpart.

    We attribute this improvement to the inclusion of structured reasoning information during pretraining, which encourages the encoder to develop a deeper understanding of the relationships between aspects and their corresponding sentiments, as well as the underlying linguistic nuances. 

    Overall, pretraining encoder models on explicit reasoning traces represents a promising direction for future research and holds potential beyond ABSA, particularly in tasks requiring fine-grained understanding and robust generalization.

    \subsection{Multilingual capabilities}
    
    With the recent advancements in multilingual models~\citep{yu2024arcticembed20multilingualretrieval, gemma_2025}, it has become increasingly important for real-world ABSA systems to support multiple languages effectively. To evaluate the multilingual capabilities of our encoder-only model, we translated the ABSA-mix dataset into six languages: English, French, German, Spanish, Italian, and Polish.

    As expected, the model trained solely on English data performs poorly on other languages. However, when trained on the multilingual version of the dataset spanning all six languages, the model demonstrates strong cross-lingual generalization without any degradation in English performance. Remarkably, it achieves consistently high results across all evaluated languages.

    To further analyze the emergence of multilingual capabilities, we conducted additional ablation studies. Table~\ref{tab:absa_multilingual_results} presents the results for models trained on English paired with each individual language, as well as on the full multilingual dataset.

    \begin{table}[h]
        \centering

        \begin{tabular}{@{}l|cccccc@{}}
            \toprule
            & \multicolumn{6}{c}{Accuracy on Language (\%)} \\
            \cmidrule(lr){2-7}
            Train & EN               & FR               & DE               & ES               & IT               & PL               \\
            \midrule
            All   & \underline{91.2} & \underline{89.4} & \textbf{88.7}    & \textbf{89.3}    & \textbf{89.0}    & \textbf{87.6}    \\
            \midrule
            EN    & \textbf{91.3}    & 64.0             & 53.2             & 69.8             & 56.0             & 40.0             \\
            +FR   & \textbf{91.3}    & \textbf{89.8}    & 71.4             & 82.9             & 79.3             & 51.6             \\
            +DE   & \underline{91.2} & 83.0             & \underline{88.0} & 82.2             & 76.8             & 61.8             \\
            +ES   & 85.3             & 56.9             & 45.4             & 81.3             & 56.0             & 37.5             \\
            +IT   & \underline{91.2} & 83.2             & 70.9             & \underline{84.1} & \underline{89.0} & 58.0             \\
            +PL   & \textbf{91.3}    & 80.6             & 66.6             & 80.5             & 74.4             & \underline{86.7} \\

            \bottomrule
        \end{tabular}
        \caption{Arctic-ABSA encoder accuracy evaluated on ABSA-mix and translations. The model was trained on pairs of English + another language and all $6$ languages. The encoder achieves strong multilingual performance without compromising accuracy in English.}
        \label{tab:absa_multilingual_results}

    \end{table}

    A particularly interesting observation is the emergence of cross-lingual capabilities even when the model is not explicitly trained on certain target languages. For example, training the model on just English and Italian significantly boosts its performance on languages such as French, German, and Polish. This highlights the model's strong ability to generalize across linguistic boundaries.

    All experiments were conducted using the same set of hyperparameters, as detailed in Section~\ref{sec:methodology}, further underscoring the robustness and generalization capacity of the approach.


    \section{Conclusion}
    In this work, we addressed several common limitations in aspect-based sentiment analysis, including the reliance on small, domain-specific datasets and a predominant focus on English-only models. We introduced a suite of decoder-only and encoder-only models that significantly outperform widely used LLMs across multiple domains, while being far more efficient and cost-effective, making them suitable for real-world deployment.

    We further demonstrated that multilingual ABSA is both feasible and effective. By translating datasets into other languages, our models retain strong performance, highlighting the potential for broad cross-lingual applicability. Additionally, our models are capable not only of performing fine-grained ABSA but also of predicting the overall sentiment of a given text, making them versatile for various sentiment analysis tasks.

    Finally, we proposed a novel reasoning injection technique for encoder-only models, which shows strong promise for pretraining more powerful encoders. We believe this approach could establish a new standard in encoder pretraining and warrants further exploration beyond the scope of ABSA tasks.

    \section{Acknowledgments}
    The authors would like to thank Mateusz Chiliński, Luke Merrick, Łukasz Borchmann, and Puxuan Yu for their valuable review and constructive suggestions on this paper. We are also very grateful to Tomasz Stanisławek, Jessie Félix, Paweł Pollak, and Łukasz Duhr for their essential engineering work and support for the project.

    \bibliography{custom}

    \section{Appendix}
    \label{sec:appendix}

               \begin{figure*}[ht!]

        \includegraphics[trim=0 18.7cm 0 0, width=\linewidth]{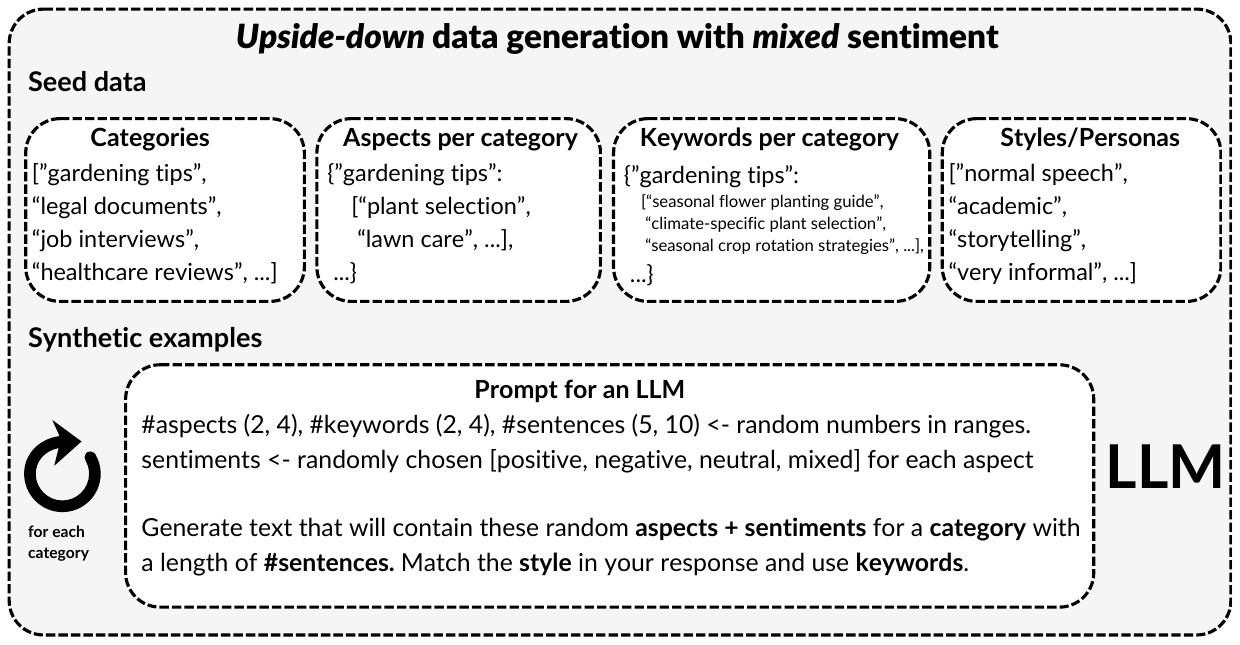}
        \caption {\textit{Upside-down} generation process of synthetic data.
        We extensively use a corpus of seed data with $46$ categories (created manually and using LLMs), $1832$ aspects, and $3463$ keywords (both generated using LLMs).
        The difference between aspects and keywords is the following: aspects are used as labels in predictions and are often general, whereas keywords only guide the generation process for better diversity and are much more descriptive.
            $18$ handcrafted styles additionally improve the diversity of the generated data.}
        \label{fig:synthetic_data}
    \end{figure*}

    \subsection{Limitations}
    Although we created a powerful collection of models for real-life aspect-based sentiment analysis, there are still some limitations.
    When analyzing model predictions, we observed that dataset labeling is often inconsistent. In particular, when expressing weak emotions, annotators frequently confuse \textit{neutral} sentiment with \textit{positive} or \textit{negative}.
    This is often due to short texts that lack broader contextual information.

    The next generation of real-life ABSA systems could greatly benefit from including more context and modalities.
    Human sentiment is inherently complex, and text may not always reflect a person's true feelings.

    Moreover, the most popular ABSA datasets typically consist only of short texts, and the performance of our model on long sequences is not guaranteed. The longest examples were in the $2048$ token range.

    \subsection{Dataset Details}
    We used a concatenation of $17$ publicly available datasets to train our models.
    All the sources with the number of samples and domains are listed in Table\ \ref{tab:dataset_sources}.
    In total, ABSA-mix spans 92 distinct domains.

    \begin{table*}[h!] 
        \centering
        \begin{tabular}{lrrrrrrrr}
            \toprule 
            \textbf{Dataset} & \textbf{\#Samples} & \textbf{\#Unique aspects} & \textbf{\#Pos} & \textbf{\#Neg} & \textbf{\#Neu} & \textbf{\#Mixed} & \textbf{\#Unk} \\
            \midrule 
            SemEval14        & 3421               & 2136                      & 3105           & 1615           & 1070           & 0                & 0              \\
            SemEval14-synth    & 3368               & 836                       & 3103           & 2090           & 3963           & 0                & 0              \\
            ABSA-public      & 63013              & 21819                     & 63949          & 34634          & 20801          & 720              & 18845          \\
            ABSA-synth       & 19503              & 1638                      & 13679          & 13022          & 13217          & 18279            & 7721           \\
            \bottomrule 
        \end{tabular}
        \caption{Statistics (number of text samples, unique aspects, and 5 sentiments) for train splits of: SemEval14 (restaurants + laptops) dataset \citep{pontiki_semeval-2014_2014}, our synthetic variant of SemEval14, selected collection of public datasets (ABSA-public), and our synthetic data (ABSA-synth). Our datasets are more than $20$ times bigger than the popular SemEval14.}
        \label{tab:dataset_stats}
    \end{table*}

    Concatenation of the datasets must be done carefully.
    Not only did we observe some leaks with the SemEval14 dataset (which we removed), but also many inconsistencies that required meticulous and time-consuming cleaning.
    Also, the differences in the distributions of data in the datasets might be pronounced in some cases as we found out that training on MAMS datasets degraded the performance of the encoder-only model on SemEval14 (restaurants).

    The difference in performance was also observed in the decoder-only model, however it was not as pronounced as in the encoder-only model.
    Therefore, when we report the results in Table\ \ref{tab:model_performance}, we do not include the MAMS datasets in the training of the encoder-only model (they are, however, included in the evaluation on ABSA-mix dataset).

    In Table\ \ref{tab:dataset_stats} we report the statistics of the datasets we created, alongside SemEval14 for comparison.
    Adding the synthetic data was very important to unlock the potential of the models to work on the \textit{mixed} and \textit{unknown} sentiments.

    \subsection{Synthetic Dataset Details}
    We created $3$ different algorithms for the creation of various synthetic datasets discussed in Section\ \ref{sec:dataset}.
    In this section, we provide more details about the algorithms.

    \subsubsection{Mixed sentiment}
    As the collection of public datasets contained only $720$ \textit{mixed} sentiment polarities (created mostly due to merging the same text with conflicting aspects), the creation of synthetic data with \textit{mixed} sentiment was crucial for the performance of our models.
    The details of our \textit{upside-down} approach are presented in Figure\ \ref{fig:synthetic_data}.

    Thanks to leveraging a \textit{seed corpus} of categories, aspects, keywords, and style personas, we could create a high-quality, very diverse dataset.
    Not only did it unlock the potential of the models to work on the \textit{mixed} sentiment, but it also increased the variety of data samples.

    When prompting the LLMs, we used few-shot prompting with CoT examples.
    Interestingly, Chain of Thought was necessary for the LLM to understand the \textit{mixed} sentiment.
    Without it, the LLM frequently generated \textit{mixed} sentiment at the overall level rather than the aspect level.
    Adding CoT examples and outputting the reasoning chain as well solved the problem and enabled the correct generation process.

    \subsubsection{Unknown sentiment}
    Algorithm\ \ref{alg:unknown_aspect} describes the process of adding the \textit{unknown} sentiment to the dataset.
    By its nature, the \textit{unknown} sentiment did not require generating new text; instead, we added \textit{unknown} aspects to existing data.
    We decided not to add the \textit{unknown} aspects to all samples and only add them to randomly selected $25\%$ of our data.

    Thanks to adding it to our synthetic dataset, we achieved a scenario where the model sees various combinations of text and aspects, which is especially important for the decoder-only model, which accepts the list of aspects as input.
    When generating the \textit{unknown} aspects, we used a random number of aspects between $2$ and $4$ (uniformly sampled) to ensure diversity.

    \begin{algorithm}[h!]
        \caption{Generate Unknown Aspect Samples}
        \label{alg:unknown_aspect}
        \begin{algorithmic}[1]
            \State \textbf{Input:} DataFrame $df$, LLM model, $p=0.25$ - probability which data points to modify, $a=2$ min number of \textit{unknown} aspects, $b=4$ max number of \textit{unknown} aspects
            \State \textbf{Output:} Modified DataFrame

            \State Calculate a set $U$ of all unique aspects from $df$ (allowed aspects)

            \For{each row in $df$}
                \State $r \sim \text{Uniform}(0, 1)$
                \State \# Modify only part of the dataframe
                \If{$r > p$}
                    \State \textbf{continue}
                \EndIf
                \\
                \State \# Select a random number of aspects
                \State $n \sim \text{Uniform}\{a, \dots, b\}$
                \State $B \gets$ Set of aspects from data point
                \State $A \gets U - B$
                \State $C \gets$ Randomly select $n$ elements from $A$

                \\
                \State \# Construct prompt using row data, $C$, $n$
                \State $prompt \gets$ \textit{create\_prompt}(row data, $C$, $n$)

                \State $result \gets LLM(prompt)$

                \If{$result$ is valid}
                    \State Add $result$ to row data
                \EndIf

            \EndFor

        \end{algorithmic}
    \end{algorithm}

    \subsubsection{SemEval14-synth}
    Algorithm\ \ref{alg:semeval_synth} describes the process of generating the synthetic data for the SemEval14 dataset.
    When generating the synthetic data, the LLM was prompted to generate samples with a similar structure as the given batch of $10$ samples while also prioritizing the \textit{neutral} sentiment.

    \begin{algorithm}[h!]
        \caption{Upscale SemEval14 dataset with synthetic data}
        \label{alg:semeval_synth}
        \begin{algorithmic}[1]
            \State \textbf{Input:} DataFrame $df$ with SemEval14 data, LLM model, $b=10$ batch size, LLM generation parameters $t=0.3$ - temperature and $top\_p=0.95$
            \State \textbf{Output:} New DataFrame

            \State new\_df = empty DataFrame
            \While{len(new\_df) < len(df)}
                \State df = shuffle(df)
                \For{i in range(0, len(df), $b$)}
                    \State batch = df[$i$:$i+b$]
                    \State prompt = \textit{create\_prompt}(batch)
                    \State result = LLM(prompt, $t$, $top\_p$)
                    \If{result is valid}
                        \State add result to new\_df
                    \EndIf
                \EndFor

            \EndWhile
            \State remove duplicates from new\_df

        \end{algorithmic}
    \end{algorithm}

    As can be seen in Table\ \ref{tab:dataset_stats}, the most common class in the SemEval14-synth dataset is in fact the \textit{neutral} sentiment with $3963$ samples compared to only $1070$ in the original SemEval14 dataset.
    By explicitly prompting the LLM to focus more on the \textit{neutral} sentiment, we were able to increase the proportions.

    We increased the temperature to $0.3$ and decreased $top\_p$ to $0.95$ to make sure the LLM generates more diverse samples.
    In all other cases, we used the temperature with a value of $0$ as we controlled the diversity with our \textit{seed corpus} and the tasks were more difficult.

    \subsection{LLM evaluation}
\begin{figure}[h!]
        \centering
        \includegraphics[trim=0 19.2cm 10.5cm 0,width=\linewidth]{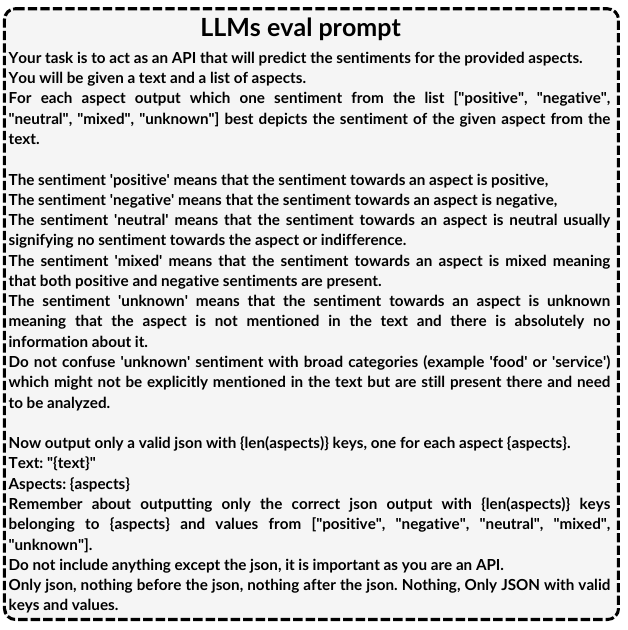}
        \caption{Prompt used for evaluation of the LLMs.
        We specify the different possible sentiments and inject \textit{text} and \textit{aspects} to be analyzed by the model.
        Additional repetitions of the JSON format improved the instruction following capabilities\ \citep{repeated_prompting}.}
        \label{fig:prompt}
    \end{figure}

    We compared our models to popular LLMs both proprietary: GPT-4o\ \citep{openai_gpt-4_2024}, Claude Sonnet 3.5\ \citep{claude35sonnet} and open-source with various sizes: Mistral Large 2\ \citep{mistral_large_2407}, Llama3.1-405b, Llama3.1-70b, Llama3.1-8b, Llama3.2-3b\ \citep{grattafiori_llama_2024}.
    In our benchmark, we used the same set of generation parameters: temperature $0$ and $top_p=1$ with the prompt presented in Figure\ \ref{fig:prompt}.

    The smallest models (8b and 3b) were accessed via vLLM\ \citep{kwon2023efficientmemorymanagementlarge} with Outlines\ \citep{willard2023efficient} backend to ensure the correct JSON output format.
    Larger models were accessed using external APIs without JSON mode as they were capable of correctly outputting the JSON format more than $99.8\%$ of the time.

    We also conducted experiments with few-shot examples (3 examples) and CoT prompting.
    The LLM results were comparable to the zero-shot setup.

    In Table\ \ref{tab:model_performance}, the LLMs were prompted using the prompt with $5$ possible output sentiments, as our models were also trained to predict $5$ possible sentiment polarities, even when the testing datasets only had $3$ possible values.
    The only exception is Table\ \ref{tab:semeval_performance} where all the models were trained and evaluated only on $3$ sentiment polarities.

    When evaluating LLMs on ABSA-mix and Overalls dataset, we slightly modified the prompt to specify that the \texttt{<overall sentiment>} aspect should be interpreted as the overall sentiment of the text.
    However, the LLMs were very robust and even without the modification in the prompt, the results were very similar, indicating good instruction-following and understanding capabilities.

    \begin{table*}[h]
        \centering
        \begin{tabular}{llr}
            \toprule
            Dataset                                                                        & Domains                                & \#samples          \\
            \midrule
            {\small eastwind/semeval-2016-absa-reviews-english-translated-stanford-alpaca} & {\small Arabic hotel reviews} & {\small 4410}               \\
            {\small Alpaca69B/reviews\_appstore\_all\_absa}                                & {\small appstore reviews}              & {\small 2082}      \\
            {\small SENTFIN}                                                               & {\small financial}                     & {\small 8554}      \\
            {\small SilvioLima/absa (DMASTE split)}                                        & {\small 8 domains}                     & {\small 6029}      \\
            {\small VocabVictor/acl2014\_absa\_twitter}                                    & {\small Twitter posts}                 & {\small 6240}      \\
            {\small jordiclive/FABSA}                                                      & {\small customer reviews}              & {\small 7573}      \\
            {\small jordiclive/OATS-ABSA}                                                  & {\small Amazon, Coursera, TripAdvisor} & {\small 3962}      \\
            {\small omymble/amazon-books-reviews-absa}                                     & {\small book reviews}                  & {\small 286}       \\
            {\small SemEval\_14\_laptops}                                                  & {\small laptops}                       & {\small 1476}      \\
            {\small SemEval\_14\_restaurants}                                              & {\small restaurants}                   & {\small 2014}      \\
            {\small SemEval\_15\_restaurants}                                              & {\small restaurants}                   & {\small 217}       \\
            {\small Ssemeval\_16\_restaurants}                                             & {\small restaurants}                   & {\small 401}       \\
            {\small siat-nlp/MAMS-for-ABSA/MAMS-ACSA}                                      & {\small Citysearch New York}           & {\small 3078}      \\
            {\small siat-nlp/MAMS-for-ABSA/MAMS-ATSA}                                      & {\small Citysearch New York}           & {\small 1695}      \\
            {\small stanfordnlp/imdb}                                                      & {\small movie reviews}                 & {\small 4998}      \\
            {\small stanfordnlp/sst2}                                                      & {\small movie reviews}                 & {\small 4998}      \\
            {\small Sp1786/multiclass-sentiment-analysis-dataset}                          & {\small various}                       & {\small 5000}      \\
            {\small SemEval14-synth}                                                         & {\small restaurants, laptops}          & {\small 3368}      \\
            {\small ABSA-synth}                                                            & {\small 46 domains}                    & {\small 19503}     \\
            \bottomrule
        \end{tabular}
        \caption{Public datasets and our synthetic data used in ABSA-mix training dataset alongside the statistics: number of text samples and the domains.
        Almost all the datasets were taken from HuggingFace. The exceptions are SemEval datasets that were downloaded from the InstructABSA GitHub repository to ensure the correct benchmarking, and FABSA downloaded from the official GitHub repository.
        A number of samples might differ from the original datasets, as we meticulously cleaned the data and merged the same text sample from different sources into one data point with one source value.}
        \label{tab:dataset_sources}
    \end{table*}

\end{document}